\documentclass{article}

    \PassOptionsToPackage{numbers, compress}{natbib}

\usepackage[preprint]{neurips_2025}
\usepackage[colorlinks=true, linkcolor=black, citecolor=black, urlcolor=blue]{hyperref}
\usepackage{booktabs}

\usepackage[utf8]{inputenc} 
\usepackage[T1]{fontenc}    
\usepackage{hyperref}       
\usepackage{url}            
\usepackage{booktabs}       
\usepackage{amsfonts}       
\usepackage{nicefrac}       
\usepackage{microtype}      
\usepackage{xcolor}         
\usepackage{graphicx}       
\usepackage{amsmath}
\usepackage{amssymb}
\usepackage{mathtools}
\usepackage{wrapfig}
\usepackage{algorithm}
\usepackage{algorithmic}
\newcommand{\MyComment}[1]{\STATE \textcolor{blue}{\texttt{// #1}}}

\title{Demystifying and Enhancing the Efficiency of Large Language Model Based Search Agents}

\author{%
  Tiannuo Yang$^{1*}$, Zebin Yao$^1$\thanks{Equal Contribution}$\:\:$, Bowen Jin$^2$, Lixiao Cui$^1$, Yusen Li$^1$, \\
  \textbf{Gang Wang$^1$, Xiaoguang Liu$^1$}\\
  $^1$ Nankai University\\
  $^2$ University of Illinois Urbana-Champaign\\
  \texttt{\{yangtn,yaozb,cuilx,liyusen,wgzwp,liuxg\}@nbjl.nankai.edu.cn}\\
  \texttt{bowenj4@illinois.edu}\\
}

\begin{document}

\maketitle

\begin{abstract}
Large Language Model (LLM)-based search agents have shown remarkable capabilities in solving complex tasks by dynamically decomposing problems and addressing them through interleaved reasoning and retrieval. However, this interleaved paradigm introduces substantial efficiency bottlenecks. First, we observe that both highly accurate and overly approximate retrieval methods degrade system efficiency: exact search incurs significant retrieval overhead, while coarse retrieval requires additional reasoning steps during generation. Second, we identify inefficiencies in system design, including improper scheduling and frequent retrieval stalls, which lead to cascading latency---where even minor delays in retrieval amplify end-to-end inference time.
To address these challenges, we introduce \texttt{SearchAgent-X}, a high-efficiency inference framework for LLM-based search agents. \texttt{SearchAgent-X} leverages high-recall approximate retrieval and incorporates two key techniques: priority-aware scheduling and non-stall retrieval. Extensive experiments demonstrate that \texttt{SearchAgent-X} consistently outperforms state-of-the-art systems such as vLLM and HNSW-based retrieval across diverse tasks, achieving up to 3.4$\times$ higher throughput and 5$\times$ lower latency, without compromising generation quality. \texttt{SearchAgent-X} is available at \url{https://github.com/tiannuo-yang/SearchAgent-X}.
\end{abstract}

\section{Introduction}

Traditional Retrieval-Augmented Generation (RAG) typically uses a sequential retrieve-then-generate approach~\cite{fan_survey_2024,gao_retrieval-augmented_2024,gupta_comprehensive_2024,huang_survey_2024,wu_retrieval-augmented_2024,yu_evaluation_2024,zhao_retrieval-augmented_2024,zhao_retrieval_2024}, which limits dynamic interaction with knowledge bases. 
Recent advancements have ushered in RAG 2.0, known as \emph{Search Agents}~\cite{trivedi2022interleaving, singh_agentic_2025,li2025search,jin_search-r1_2025,chen_research_2025,song_r1-searcher_2025,xai2024grokagents,openai2024deepresearch}. This paradigm leverages the strong reasoning capabilities of Large Language Models (LLMs), allowing for the dynamic and adaptive interleaving of reasoning steps with retrieval calls throughout the generation process. Instead of a fixed pipeline, search agents can decide \emph{when} and \emph{what} to retrieve based on LLM's ongoing reasoning, leading to significant improvements in the quality and depth of the generated responses. 
Leveraging post-training techniques similar to DeepSeek-R1, some pioneering models can even autonomously initiate retrieval actions during reasoning without intermediate supervision~\cite{jin_search-r1_2025,chen_research_2025,song_r1-searcher_2025}.

However, the improved generation quality achieved by search agents often comes at the cost of efficiency—an overhead that is nontrivial in practical deployments. In reasoning-with-search scenarios, achieving low-latency responses is critical for ensuring a seamless user experience~\cite{ray_ragserve_2024,jin_ragcache_2024}. Moreover, during post-training of LLM-based search agents, efficient model rollouts over large-scale training corpora are essential to support scalable learning.
While recent systems incorporate advanced inference optimizations—such as sequence concatenation~\cite{jin_search-r1_2025,chen_research_2025,song_r1-searcher_2025} and prefix caching~\cite{jin_ragcache_2024,vllm,zheng2024sglang}—these techniques are not specifically designed to address the unique computational challenges posed by the tight interleaving of multi-step reasoning and dynamic retrieval. 

To this end, we first conduct a systematic analysis of the efficiency factors governing LLM-based search agents, uncovering insights that diverge from the understanding of naive RAG. Our in-depth analysis reveals two key observations:
First, we demonstrate a non-monotonic relationship between retrieval accuracy and end-to-end efficiency. Both excessively high (e.g., exact search) and excessively low retrieval recall degrade overall efficiency. While aiming for perfect recall incurs unnecessary computational overhead in the retrieval phase, low recall necessitates more retrieval iterations and longer reasoning paths by the LLM to compensate (as shown in Figure~\ref{fig:retrieval_accuracy_tradeoff}). This highlights that search agent systems benefit from high-recall approximate search that effectively supports reasoning without unnecessary retrieval costs.
Second, we find that search agent systems are highly sensitive to retrieval latency. Unlike naive RAG where retrieval is largely amortized, even minor increases in retrieval time in the search agent system can cause a disproportionately large increase in end-to-end latency (Figure~\ref{fig:agentic_rag_inefficiencies}a). We attribute this magnification effect to two primary root causes: \emph{improper scheduling}, where standard policies like FCFS fail to prioritize requests that would benefit most from KV-cache reuse (Figure~\ref{fig:agentic_rag_inefficiencies}b), and \emph{retrieval stalls}, where timing misalignments between asynchronous retrieval and token generation force requests to wait, leading to unnecessary recomputation (Figure~\ref{fig:agentic_rag_inefficiencies}c).

Motivated by these findings, we propose \texttt{SearchAgent-X}, an inference system dedicated for efficient search agents. \texttt{SearchAgent-X} is designed to optimize end-to-end system throughput and latency by smoothly coordinating the interleaving of self-reasoning and retrieval. Since both overly low and high retrieval efforts lead to degraded efficiency, \texttt{SearchAgent-X} chooses to build upon a high-recall approximate retrieval method. To tackle the problem of improper scheduling, \texttt{SearchAgent-X} schedules requests with priority awareness through their real-time status to enhance KV-cache utilization. Moreover, in order to overcome frequent retrieval stalls, \texttt{SearchAgent-X} proposes a non-stall retrieval mechanism through an adaptive mechanism that allows generation to proceed without unnecessary waiting while ensuring sufficient retrieval quality.

Our extensive experiments demonstrate that \texttt{SearchAgent-X} consistently and significantly outperforms state-of-the-art baseline systems across various operational settings. In both offline and online inference scenarios, \texttt{SearchAgent-X} achieves substantial improvements in system performance (e.g., 1.3-3.4$\times$ higher throughput) by improving LLM KV-cache utilization (from 0.07 to 0.65), all while maintaining the high generation quality characteristic of search agents with exact retrieval.



\section{In-depth Analysis of LLM-based Search Agent Systems}
\subsection{Preliminary: LLM-based Search Agent Systems}
LLM-based search agent systems are designed to tackle complex requests by decomposing problems into a series of interleaved, multi-turn reasoning and information retrieval steps. This allows the LLM to adaptively seek and integrate external knowledge throughout its reasoning process. Appendix~\ref{sec:appendix-illustration} shows an example of the process of a LLM-based search agent.

\textbf{Supporting Multi-Turn Reasoning.}
Search agent systems often build on LLM inference frameworks like vLLM~\cite{jin_search-r1_2025}. They use \emph{Sequence Concatenation} for dynamic retrieval: during inference, the system monitors model output for retrieval signals. Upon such a signal, LLM decoding pauses, a query is issued, and retrieved results are concatenated with previously generated tokens to form a new, extended \emph{Sequence}. This is then re-injected into the LLM to resume reasoning.

To enhance efficiency, \emph{Prefix Cache} is commonly leveraged~\cite{jin_ragcache_2024,zheng2024sglang}. This technique stores key-value (KV) pairs from the LLM's attention mechanism for prior tokens, allowing efficient reuse in subsequent generations. This is particularly advantageous in search agents, as the concatenated sequence's prefix, excluding newly retrieved tokens, overlaps with the previous generation. Furthermore, shared system prompts across search agent requests can be cached and reused. In our evaluation, enabling prefix caching saved over 24\% of token recomputation costs.

\textbf{Sequence Scheduling.}
Efficient scheduling is vital for high throughput with limited GPU resources. Modern LLM inference frameworks utilize \emph{Iteration-Level Scheduling}, where GPU scheduling decisions occur at the single token generation step granularity~\cite{vllm,zheng2024sglang}. Compared to sequence-level scheduling~\cite{nvidia_fastertransformer,nvidia_triton_dynamic_batching}, Iteration-level scheduling avoids waiting for all sequences in a batch to complete, thus preventing bubble problems and becoming a leading solution. Frameworks like vLLM typically implement a First-Come-First-Serve (FCFS) scheduling policy per iteration.

\textbf{Retrieval Mechanism.}
On the retrieval side, semantic search techniques efficiently locate relevant external knowledge. Queries are usually encoded into dense vector representations for searching in vector space. The two primary approaches are exact nearest neighbor (ENN) search~\cite{enn} and approximate nearest neighbor (ANN) search~\cite{hnsw,scann}. Graph-based ANN methods like HNSW~\cite{hnsw} offer a favorable speed/accuracy trade-off, making them suitable for large knowledge bases.

\subsection{Investigation of Key Factors Affecting Efficiency}
Despite significant progress in high-performance LLM inference and retrieval, the LLM-based search agent's efficiency remains poorly understood. 
In this section, we analyze the influence of two key factors: 1) retrieval accuracy and 2) retrieval latency, and examine how they contribute to severe inefficiencies in current solutions. For retrieval, we assume a local search with a \textit{fixed dense encoder}.

\subsubsection{Impact of Retrieval Accuracy}

\paragraph{Insight 1:} \textit{Both overly high and overly low retrieval recall degrade end-to-end efficiency. High recall increases retrieval overhead, while low recall leads to longer reasoning steps.}

\begin{wrapfigure}{r}{0pt}
  \includegraphics[width=.43\textwidth]{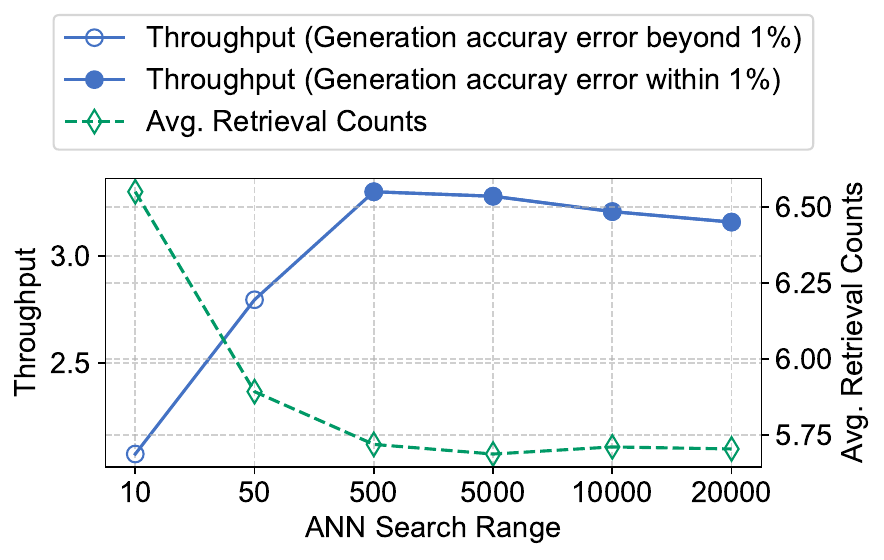}
  \caption{\textbf{Impact of Retrieval Accuracy on Search Agent Efficiency.} Higher ANN search range means higher-recall retrieval. Throughput marks the number of requests completed per second (higher is better). Retrieval count indicates the number of retrievals called per request. End-to-end generation accuracy error is calculated by comparison with an exact retrieval method. 
  }
  \label{fig:retrieval_accuracy_tradeoff}
\end{wrapfigure}

We first investigate the impact of different retrieval accuracies on the system efficiency of search agents. Intuitively, lower retrieval accuracy means lower retrieval overhead, thus higher system efficiency. However, we observe a "less is more" phenomenon for LLM-based search agents. Low-recall retrievals may result in suboptimal context, forcing the model to compensate by issuing additional retrievals and extending the reasoning length. Figure~\ref{fig:retrieval_accuracy_tradeoff} shows how varying the ANN search range affects throughput and average retrieval counts. When the search range is too small (e.g., 10), the model fails to retrieve useful documents, resulting in longer reasoning steps and an average of 6.5 retrievals per request. This reduces throughput to just above 2.1. As the search range increases to 500, retrieval quality improves, and the model completes reasoning with fewer retrievals (around 5.7), boosting throughput to over 3.2.

However, further increasing the search range (e.g., beyond 10,000) yields diminishing returns. While average retrieval counts decrease slightly, throughput declines due to the higher cost of very high-recall ANN searches. This suggests that simply maximizing retrieval recall is not the optimal strategy for search agent efficiency. Once retrieval quality sufficiently supports reasoning, additional search efforts offer marginal benefits and can even harm overall efficiency.




\begin{figure}
  \centering
  \includegraphics[width=\textwidth]{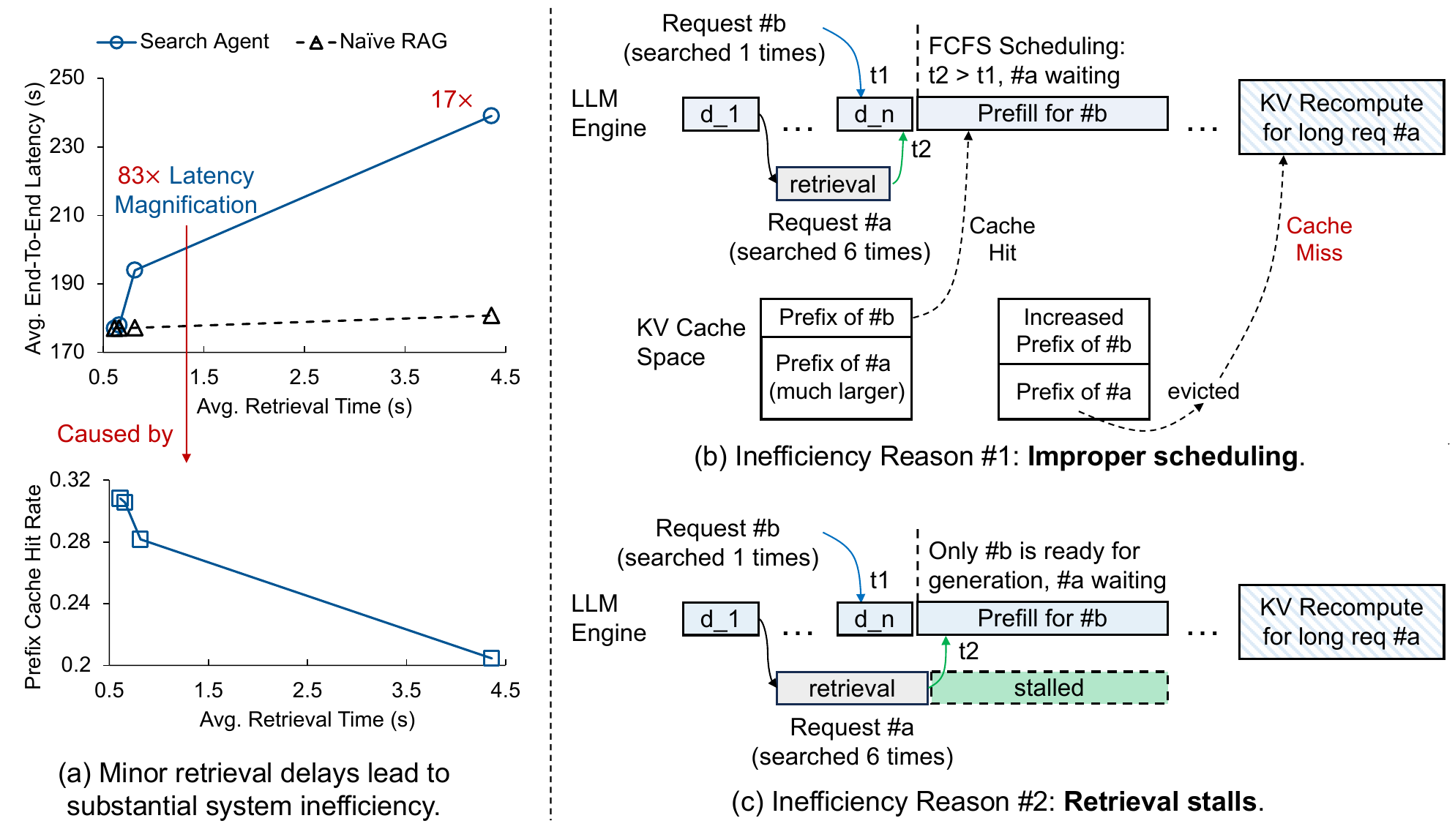}
  \caption{\textbf{Impact of Retrieval Latency on Search Agent Efficiency.} \textbf{(a)} Search agents exhibit significantly higher retrieval latency sensitivity than naive RAG (up to 83$\times$ magnification), linked to lower prefix KV-cache hit rates. \textbf{(b, c)} Root causes include: \textbf{(b)} improper scheduling, where serving shorter requests first evicts cache for longer ones, causing recomputation; and \textbf{(c)} retrieval stalls, where requests missing a scheduling point must wait for next iteration, risking cache eviction.}
  \label{fig:agentic_rag_inefficiencies}
 \end{figure}

\subsubsection{Impact of Retrieval Latency}\label{sec:retrieval_latency}






\paragraph{Insight 2:} \textit{Compared to naive RAG, search agents are much more sensitive to retrieval latency due to ignoring inter-request priorities and retrieval stalls.}

For naive RAG, all requests are retrieved before generation. Retrieval latency (millisecond level) is negligible compared to the total request latency (second level), so it is insensitive to retrieval latency. However, for search agents, retrieval occurs during self-reasoning, where the time scale of a single token generation and retrieval latency are comparable. Minor retrieval latency can cause requests to be inserted into different token generation iterations, leading to different system behaviors.

Figure~\ref{fig:agentic_rag_inefficiencies} shows the average end-to-end latency of search agents and RAG under different retrieval latency (controlled by different search ranges), with a request rate of 5 requests/second and a test duration of 10 minutes. For fair comparison, we assume RAG generates the same length of tokens with search agent, and its end-to-end latency $t^{e2e}$ is calculated as $t^{e2e}_0 + \bar{t}_{ret} \cdot \bar{n}_{ret}$, with $t^{e2e}_0$ as the token generation time without retrieval, $\bar{t}_{ret}$ as the average retrieval time, and $\bar{n}_{ret}$ as the average retrieval counts per request. The results indicate that search agents suffer from drastic efficiency degradation under even minor retrieval delays. As average retrieval latency increases from $0.6$s to $4.4$s, the end-to-end latency of the search agent is magnified by over $83\times$, while RAG remains largely stable. This severe magnification in search agent is strongly correlated with a sharp decrease in the prefix KV-cache hit rate, dropping from over $30\%$ to under $21\%$, which forces frequent and costly KV recomputations (Figure~\ref{fig:agentic_rag_inefficiencies}a).



We identify two root causes for this observed behavior, both contributing to unnecessary KV recomputation, particularly for longer, multi-turn requests: \textbf{improper scheduling} and \textbf{retrieval stalls}.
Figure~\ref{fig:agentic_rag_inefficiencies}b illustrates the issue of improper scheduling.
Consider request \#a, which involves a longer reasoning path with 6 retrievals, and request \#b, which just completes a single retrieval. Even if request \#a arrives first, if its retrieval completes slightly later than that of \#b ($t_2 > t_1$), a standard FCFS scheduler may choose to serve \#b first in the next iteration. As \#b proceeds with its generation, it occupies valuable KV-cache space, potentially leading to the eviction of the prefix KV-cache belonging to \#a. When request \#a eventually resumes, it encounters a \emph{cache miss} and must recompute its entire prefix from scratch, significantly increasing its latency. Our measurements highlight the high cost of such improper scheduling: 55.9\% of tokens were unnecessarily recomputed in affected cases, leading to more than a 108\% increase in computation time per request.

Even with improved scheduling, another significant inefficiency risk comes from reasoning stalls, depicted in Figure~\ref{fig:agentic_rag_inefficiencies}c. The asynchronous execution of retrieval and generation can lead to subtle timing misalignments. If a long request like \#a completes its retrieval only slightly after the deadline for inclusion in the next generation step, it misses the current scheduling batch and is forced to wait until the subsequent one. We term this unproductive waiting period "retrieval stalls." During this stall, shorter requests (e.g., \#b) that are ready can continue executing. Their execution may further displace \#a's prefix from the KV-cache, once again resulting in costly recomputation upon \#a's eventual resumption. Our data shows that, on average, more than 25\% of sequences experience such stalls after completing their retrieval across various scenarios.

\paragraph{Limitations of Existing Solutions.}
Our analysis highlights key limitations in current search agent systems. ENN retrieval, despite full recall, incurs prohibitive retrieval overhead. While high-recall ANN search is more suitable, it suffers from retrieval stalls due to asynchronous execution. Furthermore, prevalent FCFS scheduling in LLM inference frameworks~\cite{vllm,abhyankar_infercept_2024} disregards the search agent's unique request priorities, leading to suboptimal cache utilization and costly recomputation.


\section{Design of SearchAgent-X} \label{sec:design}
\begin{figure}
  \centering
  \includegraphics[width=\textwidth]{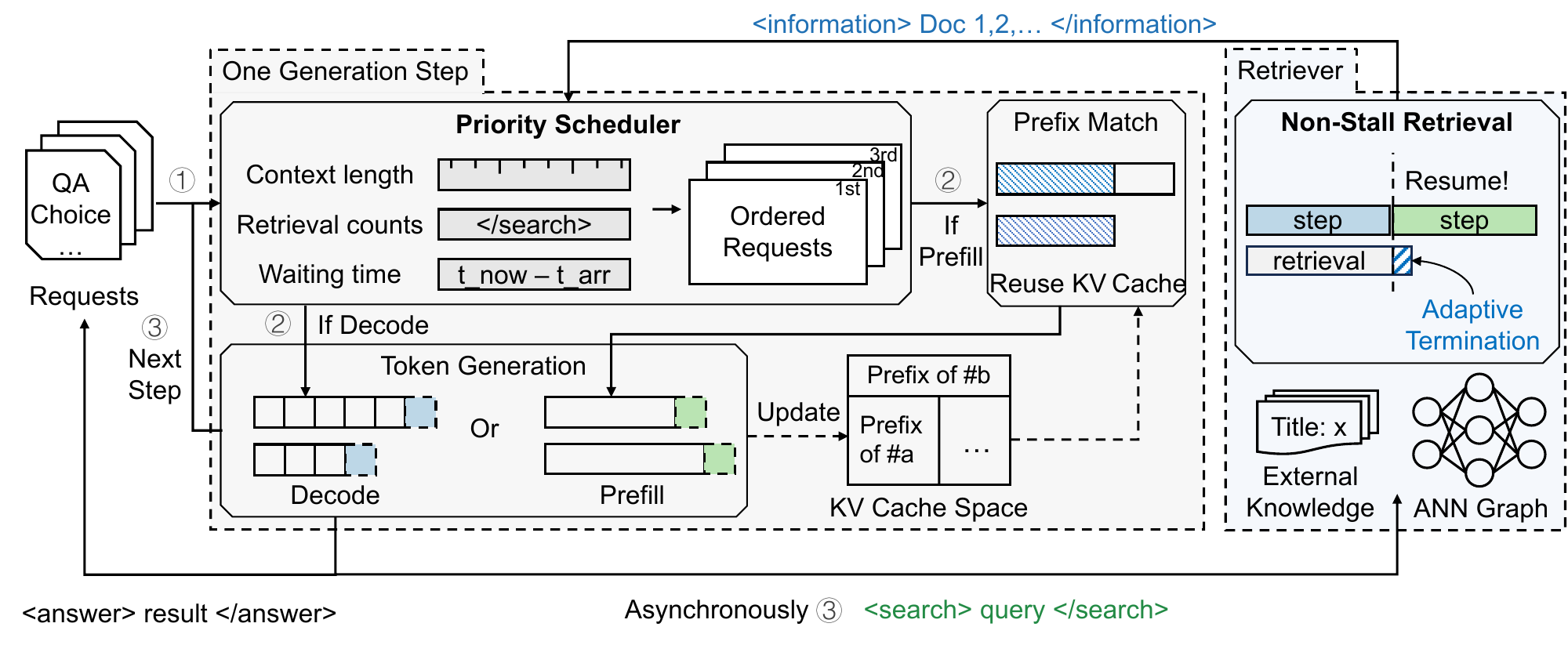}
  \caption{\textbf{\texttt{SearchAgent-X}'s Architecture.} Requests are scheduled with priorities. Reasoning and retrieval are interleaved, with a non-stall retrieval mechanism to avoid unnecessary waiting.}
  \label{fig:SearchAgent-X_architecture}
\end{figure}


\subsection{Overall Architecture} \label{sec:over_architecture}
Drawing upon the above insights, we propose \texttt{SearchAgent-X}, an inference system explicitly designed to optimize end-to-end efficiency for search agent workloads by smoothly interleaving self-reasoning and retrieval.
Figure~\ref{fig:SearchAgent-X_architecture} shows \texttt{SearchAgent-X}'s architecture, a tightly integrated system processing search agent requests at the token generation level. At each LLM output step, the system checks for special tags that trigger the Retriever for an ANN-based search (e.g., \texttt{<search>}) or request completion (e.g., \texttt{<anser>}), respectively.
To optimize GPU resource usage, \texttt{SearchAgent-X} incorporates a \textbf{priority scheduler}. It dynamically prioritizes concurrent requests using real-time collected metrics like retrieval count and waiting time, aiming to enhance KV-cache reuse by processing higher-priority requests first. During prefill, prefix matching reuses existing KV pairs from cache, significantly reducing computational overhead; new KV states are computed if caching is inapplicable or a miss occurs. Retrieval and generation operate asynchronously to enhance throughput. When retrieval is triggered, the system queries a pre-built ANN graph index. To proactively avoid retrieval stalls, \texttt{SearchAgent-X} employs a \textbf{non-stall retrieval} mechanism with adaptive search termination, which allows generation to proceed without unnecessary waiting while ensuring sufficient retrieval quality.

\subsection{Priority Scheduling} \label{sec:priority_scheduling}
\texttt{SearchAgent-X} employs a priority-based scheduling mechanism to efficiently and fairly manage concurrent generation requests. As introduced earlier, each search agent request $i$ involves a list of generation sequences $[s_{i,0}, s_{i,1}, \dots, s_{i,r_i}]$, where $s_{i,0}$ is the initial sequence and $s_{i,j}$ ($j>0$) represents a sequence resumed after the $j$-th retrieval. Let $r_i$ denote the current number of retrievals performed for request $i$, and $s_{i,r_i}$ be the sequence currently being processed.

As discussed earlier, requests that have undergone more retrieval steps (i.e., higher $r_i$) benefit more significantly from prefix cache reuse due to longer shared prefixes. Prioritizing such requests can therefore enhance overall cache efficiency and reduce redundant computation. However, scheduling solely based on retrieval count risks starving requests with fewer or no retrievals, leading to increased end-to-end latency and reduced fairness.

To mitigate these issues, \texttt{SearchAgent-X} utilizes a hierarchical scheduler that dynamically prioritizes requests based on a combination of three key metrics associated with request $i$: (1) the number of retrievals completed $R_i = r_i$; (2) the context length of the current sequence $C_i = L_{\text{seq},i}$; and (3) the waiting time of the initial request $W_i = t_{\text{now}} - t_{\text{arr},i}$. The first two metrics implicitly prioritize sequences with longer reusable prefixes, while the last ensures fairness by giving preference to requests that have been waiting longer overall.

Instead of combining these diverse metrics into a single weighted score, which would require tedious and potentially task-specific tuning of weights, \texttt{SearchAgent-X} discretizes each metric into $G$ distinct priority levels. For a given metric $M \in \{R, W, C\}$, the threshold defining the lower bound for level $k$ is calculated as:
\begin{equation}
T_{M,k} = \min(M) + \frac{k}{G} \cdot (\max(M) - \min(M)), \quad 0 \le k < G
\end{equation}
A request $i$ is then assigned to the highest priority level $k$ for which at least one of its metric values ($R_i, W_i, C_i$) exceeds the corresponding threshold $T_{M,k}$:
\begin{equation}
k = \max \left\{ j \in [0, G{-}1] \;\middle|\; R_i > T_{R,j} \lor W_i > T_{W,j} \lor C_i > T_{C,j} \right\}
\end{equation}
Requests that do not meet any threshold are assigned to the base level $0$.

Within each assigned priority level, active sequences are further sorted according to their current queueing time, defined as $W^{\text{cur}}_i = t_{\text{now}} - t_{\text{arr},i}^{(r_i)}$, where $t_{\text{arr},i}^{(r_i)}$ is the time when the sequence $s_{i,r_i}$ becomes ready for processing (e.g., after retrieval completes). Sorting by $W^{\text{cur}}_i$ in descending order ensures that among requests of similar priority level, those that have been waiting longest for their current step are processed first, mitigating the risk of KV-cache eviction during extended waits.

Finally, \texttt{SearchAgent-X} determines the execution order for a batch by traversing the priority levels from highest to lowest and processing the sequences within each level based on their sorted $W^{\text{cur}}_i$.

\subsection{Non-Stall Retrieval} \label{sec:non_stall}
To mitigate inefficiencies from retrieval latency and prevent generation stalls (Section~\ref{sec:retrieval_latency}), \texttt{SearchAgent-X} incorporates a flexible, non-stall early termination strategy for Approximate Nearest Neighbor (ANN) search. Unlike traditional ANN search that iteratively refines candidates until meeting pre-set criteria (e.g., explored nodes, list stability) and can thus cause pipeline stalls if retrieval is slow, \texttt{SearchAgent-X} adaptively concludes the search. This adaptive termination is based on two key conditions: the maturity of retrieval results and the readiness of the LLM engine, ensuring generation proceeds without unnecessary blocking.

At the core of this strategy is the concept of a \textit{soft limit} for the retrieval process. This soft limit represents a checkpoint where search results are likely to have achieved sufficient quality for the generation task. \texttt{SearchAgent-X} estimates retrieval maturity by monitoring returns in quality improvement during the ANN search. 
While retrieval quality generally improves with more explored neighbors, we find that the rate of improvement diminishes significantly after a certain point, exhibiting a "knee" where newly found points contribute less to the overall quality. \texttt{SearchAgent-X} exploits this observation. A normalized metric $\text{RQ}_t$ is used to evaluate the quality of newly discovered candidates at step $t$, defined as:
$
(d_{t} - d_{\text{best}})/(d_{\text{worst}} - d_{\text{best}}),
$
where $d_{t}$ is the new candidate's distance to the query, while $d_{\text{best}}$ and $d_{\text{worst}}$ are the distances of the best and worst candidates currently in ANN algorithm's list. 
A high $\text{RQ}_t$ value suggests the new candidate offers little improvement over existing ones, indicating diminishing returns from further search (see details in Appendix~\ref{sec:appendix-maturity}). 

The maturity exit criterion is met when this smoothed quality signal (derived from $\text{RQ}_t$) indicates a plateau (i.e., exceeds a threshold $\tau$) \textit{and} the LLM engine is ready for its next token generation operation. Upon meeting both conditions, \texttt{SearchAgent-X} halts the retrieval and provides the current, sufficiently mature results to the LLM; otherwise, retrievals stop naturally. This adaptive alignment of asynchronous retrieval and generation significantly reduces end-to-end latency without compromising the quality of the retrieved context, contrasting with traditional fixed stopping criteria that may not optimally synchronize with the dynamic state of the generation pipeline. 
\texttt{SearchAgent-X}'s complete execution process and implementation details can be found in Appendix~\ref{sec:appendix-implementation_details}.



\section{Evaluation}
\subsection{Experimental Setup}\label{sec:exp_setup}

\textbf{Models and Datasets.}
We evaluate our method using reasoning models from Search-R1~\cite{jin_search-r1_2025}, based on Qwen-7B and Qwen-14B. For retrieval, we adopt a chunked Wikipedia dataset as the knowledge base, using an ANN index constructed with HNSWlib~\cite{hnswlib}. Note that our approach is model-agnostic and readily generalizes to other reasoning models/ANN methods. 

\textbf{Testbed.}
For the 7B model, we use a single NVIDIA L20 GPU with 48GB memory. For the 14B model, we use two A100 GPUs with 40GB memory each, connected via PCIe 3.0. The retrieval system runs on 22 CPU cores with 120GB of RAM.

\textbf{Baselines.}
We compare the performance of four methods: 1) \texttt{vLLM\_ENN}: the vanilla vLLM with exact retrieval. 2) \texttt{vLLM\_ANN}: vanilla vLLM system~\cite{vllm} with approximate retrieval. 3) \texttt{CachevLLM\_ANN}: vanilla vLLM with approximate retrieval and prefix cache. 4) \texttt{SearchAgent-X}: our proposed system. Refer to Appendix~\ref{sec:appendix-setup} for the detailed experimental setup.

\subsection{End-To-End Performance}
\begin{figure}
  \centering
  \includegraphics[width=\textwidth]{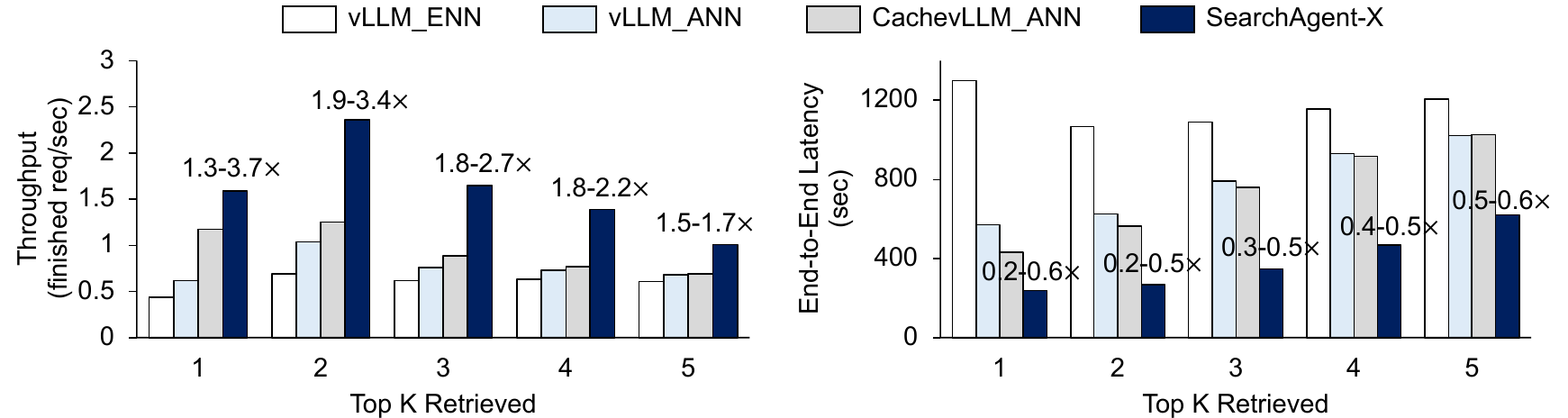}
  \caption{\textbf{End-to-End Efficiency of Offline Inference.} {Left: Requests completed per second (higher is better). Right: Average end-to-end latency (lower is better).}}
  \label{fig:efficiency_offline}
  \end{figure}


We first evaluate the end-to-end performance of different methods. For efficiency measurement, we use Musique~\cite{jin_flashrag_2025}, a dataset of complex multi-hop queries. Two scenarios are tested: (1) offline inference, where all requests arrive at the start; and (2) online inference, where requests arrive at a fixed rate. In the offline setting, we process 1000 requests and measure efficiency upon completion. In the online setting, requests arrive at rates from 1 to 6 over a 10-minute window. Results for the 7B model are shown in Figures~\ref{fig:efficiency_offline} and~\ref{fig:efficiency_online}; full results across all metrics and models are in Appendix~\ref{sec:appendix-overall-performance}.

\begin{figure}
  \includegraphics[width=\textwidth]{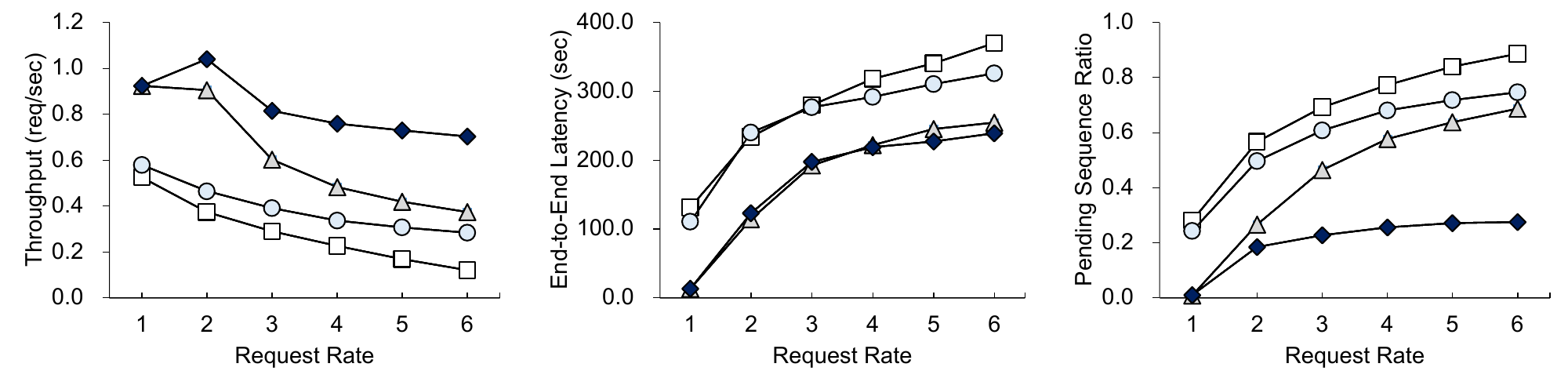}
  \caption{\textbf{End-to-End Efficiency of Online Inference.} {Left: Throughput. Middle: Latency. Right: Pending Sequence Ratio, the percentage of sequences initiated but not completed within the test period. Lower is better, indicating reasonable workload scheduling.}}
  \label{fig:efficiency_online}
\end{figure}

\textbf{In offline scenarios, \texttt{SearchAgent-X} consistently outperforms all baselines in terms of system throughput and per-request latency.} As shown in Figure~\ref{fig:efficiency_offline}, \texttt{SearchAgent-X} achieves 1.3-3.4$\times$ higher throughput and only 0.2-0.6$\times$ the latency compared to the baselines across different top-$k$ values. Even in the most challenging case of top-$k$=5, \texttt{SearchAgent-X} still beats the best baseline \texttt{CachevLLM\_ANN}, with a significant margin (1.5$\times$ in throughput and 0.6$\times$ in latency). We attribute this improvement to the \texttt{SearchAgent-X}'s high-recall ANN, and the design of efficient scheduling and non-stall retrieval mechanisms. \texttt{vLLM\_ENN} performs poorly in this scenario, as it incurs excessive retrieval overhead and hinders end-to-end reasoning efficiency. \texttt{vLLM\_ANN} employs a high-recall ANN and performs obviously better than \texttt{vLLM\_ENN}, but it still suffers from the inefficiencies of large amounts of recomputation due to the lack of prefix cache. \texttt{CachevLLM\_ANN} uses prefix cache to reduce recomputation, but it still fails to wisely manage the scheduling of requests and avoid retrieval stalls, leading to a significant performance gap compared to \texttt{SearchAgent-X}.

We also find that the performance of all methods first increases then decreases with the increase of top-$k$ values. This aligns well with our previous observation that both overly high and overly low retrieval quality degrade end-to-end efficiency. When the top-$k$ value is too small, the model may fail to retrieve useful documents, leading to longer reasoning sequences and lower throughput. Conversely, when the top-$k$ value is too large, the concatenated sequence becomes too long, resulting in longer prefill time. However, we note that \texttt{SearchAgent-X} consistently outperforms all baselines across all top-$k$ values, indicating its robustness to different retrieval settings.

\textbf{In online scenarios, \texttt{SearchAgent-X} utilizes computing resources more efficiently than baselines, completing more requests in the same amount of time.} As shown in Figure~\ref{fig:efficiency_online}, \texttt{SearchAgent-X} completes at least 1.5$\times$, and up to 3.5$\times$ more requests on average than the baseline, within the request rate range of 1 to 6. We also observe that the advantage of \texttt{SearchAgent-X} over the baselines increases with the request rate. For example, at a request rate of 6, \texttt{SearchAgent-X} achieves 5.8$\times$ more requests than \texttt{vLLM\_ENN}, and 1.9$\times$ more than the most competitive baseline \texttt{CachevLLM\_ANN}. This is because high request rates mean more contention for GPU resources across requests, while \texttt{SearchAgent-X}'s priority scheduling and non-stall retrieval significantly improve KV-cache utilization and reduce recomputation, thus mitigating resource contention. Besides, we observe that the latency of methods with prefix cache (\texttt{CachevLLM\_ANN} and \texttt{SearchAgent-X}) is obviously lower than \texttt{vLLM\_ENN} and \texttt{vLLM\_ANN}, indicating prefix cache's benefit of reducing prefill time. \texttt{CachevLLM\_ANN} incurs similar latency as \texttt{SearchAgent-X}, because it only consumes half of the requests as \texttt{SearchAgent-X}. Further, we record the pending sequence ratio that measures the resource utility of the system, defined as the percentage of sequences that are initiated but not completed within the test period. As shown in Figure~\ref{fig:efficiency_online} (right), \texttt{SearchAgent-X} achieves stable, small pending sequence ratios (about 0.2), while the baselines experiences dramatic increases with higher request rates (more than 0.6), indicating the effectiveness of \texttt{SearchAgent-X}'s scheduling.

\begin{table}
  \centering
  \small
  \setlength{\tabcolsep}{5pt}
  \renewcommand{\arraystretch}{1.2}
  \caption{Generation Quality of \texttt{SearchAgent-X} and Exact Retrieval.}
  \begin{tabular}{lccccccc}
    \toprule
    \textbf{Dataset} & Musique & NQ & 2Wiki & HotpotQA & Bamboogle & StrategyQA & Avg. \\
    \midrule
    \textbf{Generation Accuracy} & & & & & & & \\
    \quad Exact Retrieval & 0.203 & 0.316 & 0.371 & 0.429 & 0.472 & 0.788 & 0.430 \\
    \quad \texttt{SearchAgent-X}       & 0.203 & 0.320 & 0.370 & 0.428 & 0.472 & 0.789 & 0.430 \\
    \midrule
    \textbf{Retrieval Counts} & & & & & & & \\
    \quad Exact Retrieval & 3.247 & 2.288 & 3.126 & 2.702 & 2.440 & 2.496 & 2.717 \\
    \quad \texttt{SearchAgent-X}       & 3.251 & 2.292 & 3.138 & 2.699 & 2.448 & 2.476 & 2.717 \\
    \midrule
    \textbf{Output Length} & & & & & & & \\
    \quad Exact Retrieval & 8125 & 5839 & 7575 & 6840 & 6152 & 6402 & 6822 \\
    \quad \texttt{SearchAgent-X}       & 8134 & 5847 & 7600 & 6839 & 6151 & 6382 & 6826 \\
    \bottomrule
  \end{tabular}
  \label{tab:SearchAgent-X-comparison}
\end{table}

\textbf{\texttt{SearchAgent-X} achieves comparable generation quality to exact retrieval.} We evaluate the generation quality of \texttt{SearchAgent-X} and exact retrieval (\texttt{vLLM\_ENN}) on six representative datasets. We use the \emph{Exact Match} metric as generation accuracy to measure the correctness of the generated answers~\cite{jin_flashrag_2025}. As shown in Table~\ref{tab:SearchAgent-X-comparison}, \texttt{SearchAgent-X} achieves similar generation accuracy, retrieval counts and output length as exact retrieval across all datasets, indicating that it does not compromise generation quality for efficiency. Another interesting finding is that \texttt{SearchAgent-X} may even achieve higher generation accuracy on some datasets, such as NQ (0.320 vs. 0.316). The results could be attributed to two aspects. First, full ANN recall does not necessarily mean optimal generation accuracy; the correct answer document may not always be captured by semantic similarity. Second, the search agent model has the adaptability to adjust the reasoning length. ANN might lead search agents to perform an extra reasoning step (e.g., 2.292 vs. 2.288 for NQ), adjusting the retrieval query, which in turn improves generation accuracy.

\subsection{Ablation Study}
\begin{figure}
  \centering
  \includegraphics[width=\textwidth]{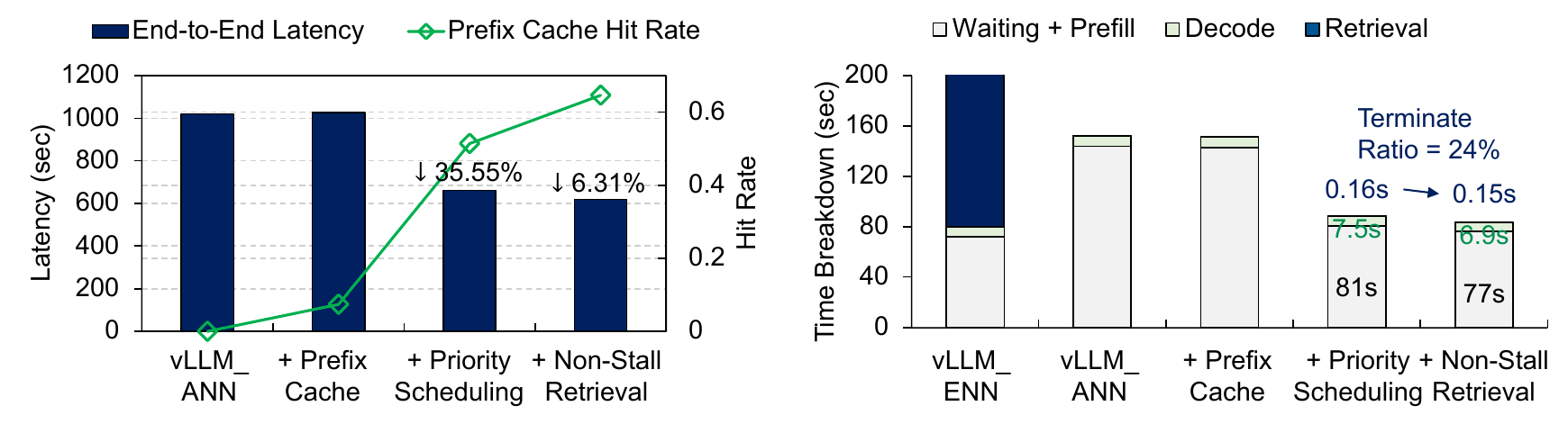}
  \caption{\textbf{\texttt{SearchAgent-X} Technique Breakdown for End-to-End Performance (Left) and Per-Sequence Generation Time (Right).} {Each bar/scatter adds one technique over its left bar/scatter, with the leftmost being vanilla vLLM and the rightmost being the full \texttt{SearchAgent-X}.}}
  \label{fig:SearchAgent-X_technique_breakdown}
\end{figure}


\textbf{The priority scheduling and non-stall retrieval of \texttt{SearchAgent-X} help improve the prefix cache utility, thus enhancing end-to-end efficiency.} Figure~\ref{fig:SearchAgent-X_technique_breakdown} (left) shows the end-to-end performance of different techniques for offline inference with top-$k$ = 5. We have several observations. First, the advantages of prefix cache are diminished in this challenging scenario. With top-$k$=5, it's only 1.01$\times$ that of \texttt{vLLM\_ANN}, compared to 1.91$\times$ with top-$k$=1. This validates that the benefits of prefix cache still require appropriate scheduling and retrieval methods to unleash its potential. Second, \texttt{SearchAgent-X}'s priority scheduling reduces the end-to-end latency by 35.55\% based on prefix cache. This is because the priority of requests is properly managed, maximizing the utilization of GPU resources. In addition, the prefix cache hit rate increases from 0.07 to 0.51, verifying the effectiveness of the technique. Third, \texttt{SearchAgent-X}'s non-stall retrieval further improves the hit rate to 0.65, leading to a further 6.3\% reduction in latency. This shows that the adaptive termination strategy fully utilizes the "free lunch" of asynchronous execution, timely recalling mature retrieval results, thereby improving system processing efficiency.

Figure~\ref{fig:SearchAgent-X_technique_breakdown} (right) further demonstrates the per-sequence generation time of different parts. We have more observations. First, for \texttt{vLLM\_ENN}, the retrieval time is the largest component, accounting for over 60\% of the total time. Instead, its prefill time is the lowest across different techniques, since its reasoning requires waiting for long-time retrieval, thus reducing the pressure on token generation.
Second, for priority scheduling, we note that it reduces not only the prefill time (due to more prefix cache utilized), but also the decode time, showing a better system processing capability. This is because by improving KV-cache utilization, it avoids recomputation of long requests, freeing up GPU space earlier for better decode parallelism.
Third, non-stall retrieval actually only reduces 0.01s of retrieval time (from 0.16s to 0.15s), with about 24\% of the retrievals being early terminated, but significantly reduces the end-to-end latency (41s). This aligns well with our previous observation that minor retrieval latency can cause drastic efficiency degradation (as shown in Figure~\ref{fig:agentic_rag_inefficiencies}). Non-stall retrieval adaptively terminates only a small set of retrievals when necessary, yet yields the significant benefit of better cache utilization. 
More ablation results can be found in Appendix~\ref{sec:appendix-priority-level} and~\ref{sec:appendix-analysis-maturity-exit}.  


\section{Related Work}
Prior work on end-to-end RAG efficiency—spanning caching~\cite{jin_ragcache_2024, lu_turborag_2024, yao2025cacheblend}, pipelining~\cite{jiang2025piperag, yu_aquapipe_2025}, and hyperparameter tuning~\cite{ray_ragserve_2024,fu_autorag-hp_2024}—primarily targets the traditional sequential “retrieve-then-generate” paradigm. These methods are generally ill-suited for tightly interleaved, multi-turn reasoning and dynamic retrievals characteristic of search agents. Meanwhile, broader agent workflow optimizations, such as auto-tuning~\cite{he2025cognify}, KV-cache management~\cite{abhyankar_infercept_2024, shahout2024fast}, and partial tool execution~\cite{xu2024conveyor}, improve overall efficiency but overlook the specific challenges of retrieval accuracy and latency in search agents. In contrast, \texttt{SearchAgent-X} directly addresses these challenges by tightly coupling priority-scheduled reasoning with non-stall retrieval, yielding improved efficiency. Notably, our approach is orthogonal to, and can potentially be combined with, other RAG optimization techniques such as context compression~\cite{shi2024compressing}, hardware acceleration~\cite{quinn_accelerating_2024}, and retrieval reranking~\cite{glass2022re2g}.

\section{Conclusion}
LLM reasoning-driven search agents offer great potential for complex problems, but face severe, distinct efficiency challenges. This paper highlights the non-trivial impact of retrieval accuracy and the severe latency sensitivity caused by scheduling deficiencies and retrieval stalls. Our proposed system, \texttt{SearchAgent-X}, designed on these insights, demonstrates substantial improvement in system efficiency, all while maintaining high generation quality. 
This study provides important insights for practical deployments of high-efficiency LLM-based search agents, and the proposed solutions are easily adaptable to other ANN retrieval methods and LLM reasoning models.


{
\small
\bibliographystyle{nips}
\bibliography{neurips_2025}
}

\clearpage
\appendix
\section{An Illustration of LLM-Based Search Agents} \label{sec:appendix-illustration}
\begin{figure}
  \centering
  \includegraphics[width=\textwidth]{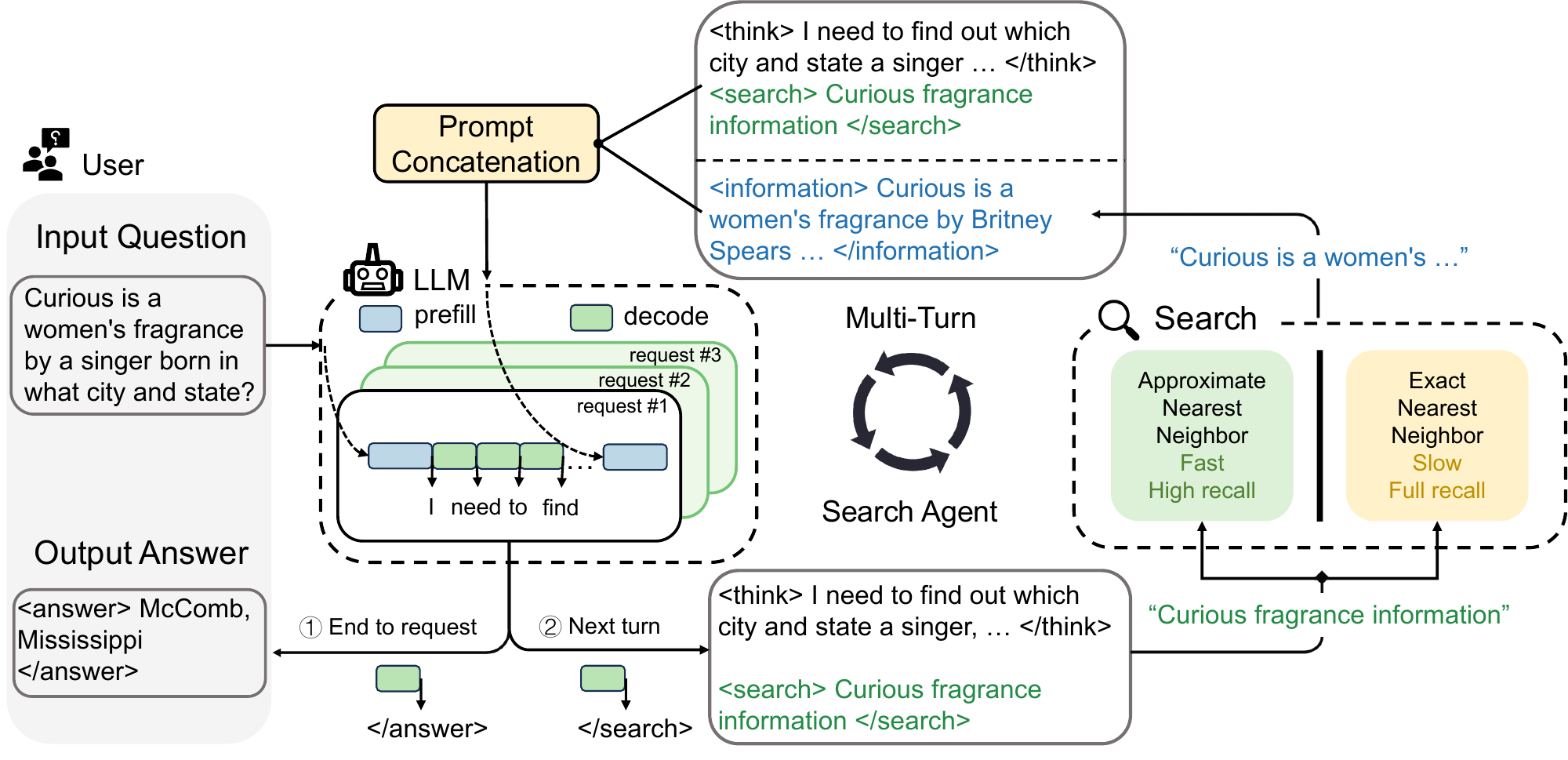}
  \caption{An illustration of reasoning and search interleaved LLM-based search agents. 
  }
  \label{fig:agentic_rag_architecture}
\end{figure}

Figure~\ref{fig:agentic_rag_architecture} shows an example of a search agent process. Faced with a complex query ("Curious is a women's fragrance by a singer born in what city and state?"), the search agent first engages in preliminary reasoning ("I need to find out which city and state a singer..."). Recognizing a knowledge gap regarding the "Curious fragrance," the model proactively decides to initiate a search ("search Curious fragrance information"). Upon receiving the crucial information ("Curious is a women's fragrance by Britney Spears"), the model doesn't conclude its process. Instead, it integrates this new knowledge into its subsequent thought process and reasoning. This triggers further searches, of which the retrieval result is concatenated with previously generated tokens and re-injected into LLMs. Through this dynamic cycle of "think-search-rethink," the model progressively assembles the necessary pieces of the knowledge puzzle required to answer the question fully. This culminates in a high-quality answer that addresses all aspects of the initial query ("McComb, Mississippi"). This ability to autonomously plan retrieval actions and iteratively incorporate new information into its reasoning process allows the search agent to tackle more complex questions and deliver better responses, moving beyond reliance solely on pre-trained knowledge or a single retrieval.

\section{Implementation Details} \label{sec:appendix-implementation_details}
\subsection{SearchAgent-X Execution} \label{sec:appendix-execution}
This section outlines the high-level execution flow of the \texttt{SearchAgent-X} system, as depicted in Algorithm~\ref{alg:SearchAgent-X_main_loop_appendix}, complementing the conceptual component descriptions in Section~\ref{sec:design} of the main paper. \texttt{SearchAgent-X} orchestrates LLM inference (with prefix caching, Section~\ref{sec:over_architecture}), dynamic high-recall approximate retrieval, Priority Scheduling (Section~\ref{sec:priority_scheduling}), and Non-Stall Retrieval (Section~\ref{sec:non_stall}) to achieve efficient search agents. The system initializes an LLM inference engine and manages incoming requests, active asynchronous search tasks, and their results. 

The main execution loop begins by ingesting new user requests into the LLM engine's pool (line5-7). Concurrently, if Non-Stall Retrieval is active, \texttt{SearchAgent-X} consults an external signal to identify and expedite the completion of any ongoing retrieval tasks that have reached sufficient maturity or for which LLM engine readiness dictates early termination (line 10-11), thus preventing pipeline stalls. Upon completion of a search (either normally or via early termination), retrieved documents are concatenated with the original context, and the augmented request is resubmitted to the LLM engine (line 15-20).

The core of the processing loop involves LLM generation and the agentic control flow. Before each LLM generation step, \texttt{SearchAgent-X}'s Priority Scheduling policy is applied to the queue of waiting requests, reordering them to optimize system throughput and KV-cache utilization (line 23). Following token generation by the LLM, the output for each active sequence is parsed (line 25-28). If a \texttt{<search>} tag is detected, indicating a need for external knowledge, \texttt{SearchAgent-X} halts further generation for that sequence and launches an asynchronous high-recall ANN retrieval task (line 29-34). Conversely, if a \texttt{<answer>} tag is identified or the sequence naturally concludes, the request is finalized (line 35-38). This iterative and asynchronous process enables the dynamic interleaving of LLM reasoning, external knowledge retrieval, and intelligent scheduling, which is fundamental to \texttt{SearchAgent-X}'s efficient handling of complex search agent workloads.


\begin{algorithm}
  \caption{\texttt{SearchAgent-X} Main Execution Loop}
  \label{alg:SearchAgent-X_main_loop_appendix}
  \begin{algorithmic}[1]
  \STATE Initialize LLM\_Engine, ArrivalQueue, ActiveSearchTasks, FinishedOutputs
  \STATE Configure PriorityScheduling (enabled/type), NonStallRetrieval (enabled)
  
  \WHILE{LLM\_Engine has unfinished requests \OR \NOT ActiveSearchTasks is empty \OR \NOT ArrivalQueue is empty}
      \MyComment{Step 1: Ingest new requests}
      \FOR{each request $R_{new}$ in ArrivalQueue ready for processing}
          \STATE Add $R_{new}$ to LLM\_Engine's request pool
          \STATE Remove $R_{new}$ from ArrivalQueue
      \ENDFOR
  
      \MyComment{Step 2: Non-Stall Retrieval Check (if enabled)}
      \IF{NonStallRetrieval is enabled \AND ActiveSearchTasks is not empty \AND LLM\_Engine has waiting requests}
          \STATE TerminatedSearchIDs $\leftarrow$ CheckExternalNonStallSignal() \MyComment{Queries for searches to terminate early}
      \ENDIF
  
      \MyComment{Step 3: Process completed search tasks}
      \FOR{each search task $S_i$ in ActiveSearchTasks that has completed}
          \STATE $R_{orig}, \text{retrieved\_docs}, \text{search\_finish\_time} \leftarrow S_i.\text{getResult()}$
          \STATE $\text{new\_context} \leftarrow \text{Concatenate}(R_{orig}.\text{context}, \text{retrieved\_docs})$
          \STATE AddResumedRequest($R_{orig}$, new\_context, search\_finish\_time) to LLM\_Engine
          \STATE Remove $S_i$ from ActiveSearchTasks
      \ENDFOR
  
      \MyComment{Step 4: LLM Generation Step}
      \IF{LLM\_Engine has unfinished requests}
          \MyComment{Section~\ref{sec:priority_scheduling}}
          \STATE ApplyPriorityScheduling(LLM\_Engine.waiting\_requests) 
          \STATE LLM\_Outputs, Scheduler\_Info $\leftarrow$ LLM\_Engine.step()
          \STATE RecordTokenTimingsAndPrefixCacheStats(Scheduler\_Info)
          \FOR{each output $O_j$ in LLM\_Outputs}
              \STATE $\text{current\_text} \leftarrow O_j.\text{getGeneratedText()}$
              \IF{DetectSearchTag($\text{current\_text}$)}
                  \STATE $\text{query} \leftarrow \text{ExtractSearchQuery}(\text{current\_text})$
                  \STATE LLM\_Engine.abortRequest($O_j.\text{request\_id}$)
                  \STATE $S_{new} \leftarrow$ LaunchAsyncRetrievalTask($O_j.\text{request\_id}$, $\text{current\_text}$, query) \MyComment{High-recall ANN}
                  \STATE Add $S_{new}$ to ActiveSearchTasks
              \ELSIF{DetectAnswerTag($\text{current\_text}$) \OR $O_j.\text{isFinished()}$}
                  \STATE Add $O_j$ to FinishedOutputs
                  \STATE LLM\_Engine.abortRequest($O_j.\text{request\_id}$)
              \ENDIF
          \ENDFOR
      \ENDIF
  \ENDWHILE
  \STATE CleanupRemainingTasks()
  \end{algorithmic}
  \end{algorithm}

\subsection{Retrieval Maturity Estimation} \label{sec:appendix-maturity}
\begin{figure}
    \centering
    \begin{minipage}[t]{0.48\textwidth}
        \centering
        \includegraphics[width=\linewidth]{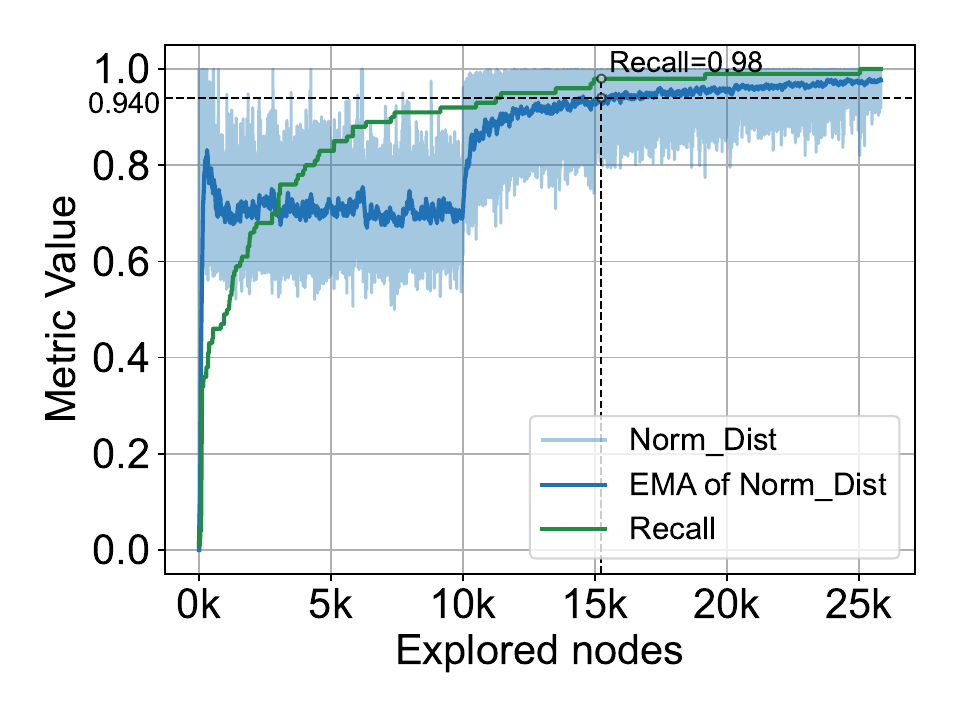}
        \caption{\textbf{EMA Signal of Retrieval Maturity.} {
        RQ means relative quality of newly explored candidates. EMA represents smoothed RQ. The vertical line marks maturity, where the improvement of recall and EMA decreases to about 0.}}
        \label{fig:ema-recall}
    \end{minipage}
    \hfill
    \begin{minipage}[t]{0.48\textwidth}
        \centering
        \includegraphics[width=\linewidth]{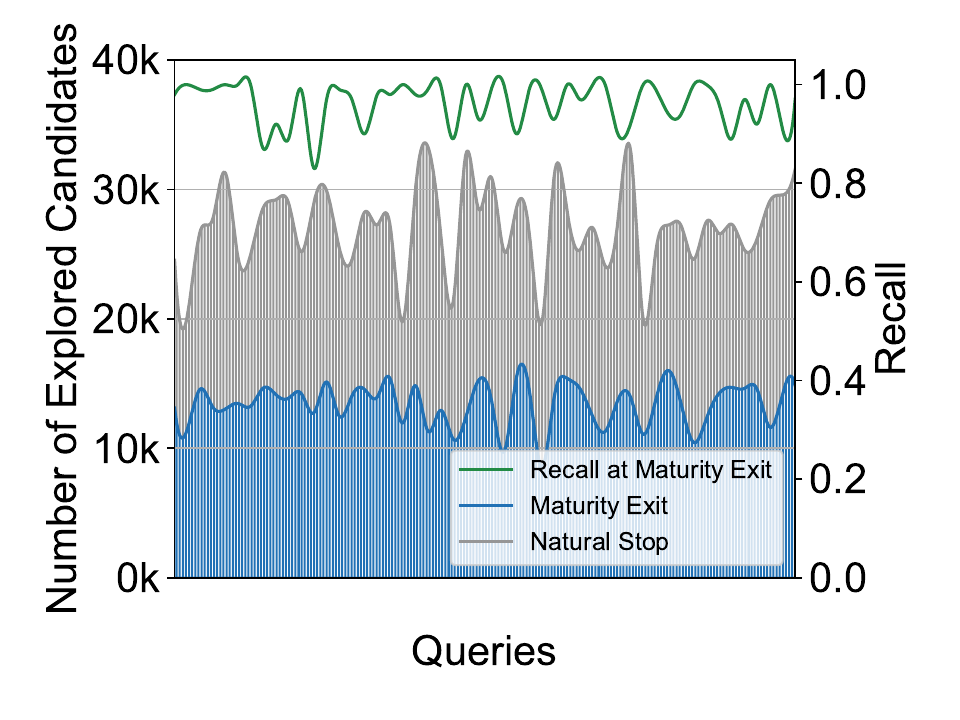}
        \caption{\textbf{Comparison of Maturity Exit and Natural Stop.} {The shadows represent the number of candidates explored by two methods (showing a similar trend). The curve represents the recall of maturity exit (consistently high).}}
        \label{fig:node-recall}
    \end{minipage}
\end{figure}

In practice, the raw $\text{RQ}_t$ signal described in Section~\ref{sec:non_stall} exhibits short-term fluctuations that may hinder robust maturity estimation. To address this, \texttt{SearchAgent-X} applies an exponential moving average (EMA)~\cite{ema}, with a window size of 500, to smooth the signal.

Selecting an appropriate threshold $\tau$ for the EMA is critical to balancing retrieval quality and latency. To determine a suitable value, we sample queries from the Musique dataset and record the evolution of the EMA curve as the number of explored candidates increases. For each query, we identify the point where the EMA curve flattens—i.e., where marginal improvements approach zero—indicating that newly explored candidates are far from the query and contribute little to quality. This point reflects the onset of retrieval maturity. As shown in Figure~\ref{fig:ema-recall}, the recall at this stage also stabilizes and reaches a high level (around 0.98). We adopt the corresponding EMA value ($\tau = 0.9$) at this "knee" as the practical threshold $\tau$ in \texttt{SearchAgent-X} to reliably trigger maturity exit.

\subsection{Detailed Experimental Setup} \label{sec:appendix-setup}
We implement \texttt{SearchAgent-X} by building upon vLLM \cite{vllm}, a state-of-the-art LLM inference engine to use its efficient PagedAttention mechanism.  For retrieval component, we use a knowledge base constructed from a chunked Wikipedia dataset, containing approximately 21 million text chunks. Each chunk is embedded into a 384-dimensional vector using the \texttt{all-MiniLM-L6-v2} model~\cite{sentencetransformer}. An Approximate Nearest Neighbor (ANN) index is built offline over these embeddings using HNSWlib~\cite{hnswlib}, configured with parameters such as up to 32 neighbors per node and an efConstruction (candidate list size during build) of 500. This index serves as the foundation for all ANN-based retrieval methods in our experiments. For these ANN searches (employed by \texttt{SearchAgent-X} and approximate retrieval baselines), we generally set the search range (e.g., efSearch) to 10,000 to achieve high recall with acceptable computational overhead, based on empirical analysis. These HNSW ANN searches leverage inter-query parallelism with 4 threads to optimize throughput while managing memory access contention.

Specific configurations for the different systems are as follows.
For the exact retrieval baseline (\texttt{vLLM\_ENN}), we adapt HNSWlib to perform exhaustive search more efficiently by enabling intra-query parallelism, utilizing 6 threads.
For \texttt{SearchAgent-X}, beyond using the aforementioned high-recall ANN setup, we set its unique parameters: the priority scheduling level $G$ is configured to $6$ (we note that \texttt{SearchAgent-X} exhibits low sensitivity to this specific value, as shown in our ablation study in Appendix~\ref{sec:appendix-priority-level}). The threshold $\tau_{EMA}$ for estimating retrieval maturity in the non-stall mechanism is set to 0.9, determined via offline profiling detailed in Appendix~\ref{sec:appendix-maturity}. The approximate retrieval baselines (\texttt{vLLM\_ANN}, \texttt{CachevLLM\_ANN}) also utilize the general ANN search settings described above, including the search range of 10,000 and 4 threads for inter-query parallelism.

\section{More Results} \label{sec:appendix-more_results}
\begin{table}
  \centering
  \caption{Comparison across seven key metrics and top-$k$ values for different methods. Throughput and efficiency gains are marked by $\times$ multipliers. Lower values are better for metrics marked with $\downarrow$.}
  \small
  \renewcommand{\arraystretch}{1.2}
  \setlength{\tabcolsep}{5pt}
  \begin{tabular}{lccccc|ccc}
  \toprule
  \textbf{Metric}
  & \textbf{Top-1} & \textbf{Top-2} & \textbf{Top-3} & \textbf{Top-4} & \textbf{Top-5} & \textbf{Top-1} & \textbf{Top-2}& \textbf{Top-3}\\
   & \multicolumn{5}{c|}{\textbf{Qwen-7B}} & \multicolumn{3}{c}{\textbf{Qwen-14B}} \\
  
  \midrule
  \textbf{Throughput} & & & & & \\
  \quad \texttt{vLLM\_ENN}      & 0.44 & 0.69 & 0.62 & 0.63 & 0.61 & 0.46 & 0.47 & 0.43\\
  \quad \texttt{vLLM\_ANN}      & 0.62 & 1.04 & 0.76 & 0.73 & 0.68 & 0.94 & 0.77 & 0.61\\
  \quad \texttt{CachevLLM\_ANN} & 1.18 & 1.25 & 0.89 & 0.77 & 0.69 & 1.08 & 0.89 & 0.70\\
  \quad \texttt{SearchAgent-X}      & 1.59 & 2.36 & 1.64 & 1.39 & 1.01 & 1.40 & 1.09 & 0.76\\
  \quad Max Ratio      & $3.61\times$ & $3.42\times$ & $2.65\times$ & $2.20\times$ & $1.66\times$ & $3.04\times$ & $2.32\times$ & $1.77\times$\\
  \quad Min Ratio      & $1.35\times$ & $1.89\times$ & $1.84\times$ & $1.81\times$ & $1.46\times$ & $1.30\times$ & $1.22\times$ &  $1.09\times$\\
  \midrule
  \textbf{Token Throughput} & & & & & \\
  \quad \texttt{vLLM\_ENN}      & 69.90 & 90.85 & 86.12 & 97.79 & 84.26 & 156.35 & 127.28 & 111.21\\
  \quad \texttt{vLLM\_ANN}      & 101.21 & 136.81 & 106.09 & 114.28 & 94.65 & 320.76 & 206.46 & 159.70\\
  \quad \texttt{CachevLLM\_ANN} & 191.65 & 164.46 & 124.59 & 119.17 & 95.64 & 366.85 & 239.01 & 182.60\\
  \quad \texttt{SearchAgent-X}      & 259.94 & 309.73 & 229.96 & 216.21 & 139.25 & 472.76 & 292.13 & 199.14\\
  \quad Max Ratio      & $3.72\times$ & $3.41\times$ & $2.67\times$ & $2.21\times$ & $1.65\times$ & $3.02\times$ & $2.30\times$ & $1.79\times$\\
  \quad Min Ratio      & $1.36\times$ & $1.88\times$ & $1.85\times$ & $1.81\times$ & $1.46\times$ & $1.29\times$ & $1.22\times$ & $1.09\times$\\
  \midrule
  \textbf{Latency} $\downarrow$ & & & & & \\
  \quad \texttt{vLLM\_ENN}      & 1300.56 & 1066.05 & 1089.37 & 1154.62 & 1205.16 & 1642.86 & 1567.85 & 1614.71\\
  \quad \texttt{vLLM\_ANN}      & 571.36 & 625.33 & 790.68 & 930.29 & 1020.46 & 767.46 & 923.74 & 1052.74\\
  \quad \texttt{CachevLLM\_ANN} & 429.60 & 562.47 & 759.57 & 916.58 & 1026.91 & 673.35 & 805.60 & 980.35\\
  \quad \texttt{SearchAgent-X}      & 238.00 & 266.50 & 347.14 & 466.78 & 620.07 & 502.10 & 690.20 & 939.42\\
  \quad Max Ratio  & $0.18\times$ & $0.25\times$ & $0.32\times$ & $0.40\times$ & $0.51\times$ & $0.31\times$ & $0.44\times$ & $0.58\times$\\
  \quad Min Ratio  & $0.55\times$ & $0.47\times$ & $0.46\times$ & $0.51\times$ & $0.60\times$ & $0.75\times$ & $0.86\times$ & $0.96\times$ \\
  \midrule
  \textbf{P99 Latency} $\downarrow$ & & & & & \\
  \quad \texttt{vLLM\_ENN}      & 1758.64 & 1441.18 & 1462.45 & 1569.80 & 1641.90 & 2205.26 & 2136.75 & 2237.55\\
  \quad \texttt{vLLM\_ANN}      & 915.04 & 956.46 & 1160.86 & 1348.87 & 1459.13 & 1066.63 & 1334.67 & 1555.47\\
  \quad \texttt{CachevLLM\_ANN} & 609.70 & 797.46 & 1073.03 & 1296.03 & 1446.88 & 930.06 & 1159.02 & 1454.61\\
  \quad \texttt{SearchAgent-X}      & 373.20 & 421.32 & 566.28 & 716.88 & 993.98 & 732.16 & 958.78 & 1362.24\\
  \quad Max Ratio  & $0.21\times$ & $0.29\times$ & $0.39\times$ & $0.46\times$ & $0.61\times$ & $0.33\times$ & $0.45\times$ & $0.61\times$ \\
  \quad Min Ratio  & $0.61\times$ & $0.53\times$ & $0.53\times$ & $0.55\times$ & $0.69\times$ & $0.79\times$ & $0.83\times$ & $0.94\times$\\
  \bottomrule
  \end{tabular}

  \label{tab:SearchAgent-X-all-metrics-2}
\end{table}

\begin{table}
  \centering
  \small
  \setlength{\tabcolsep}{5pt}
  \renewcommand{\arraystretch}{1.2}
  \caption{Generation Quality of \texttt{SearchAgent-X} and Exact Retrieval.}
  \begin{tabular}{lccccccc}
    \toprule
    \textbf{Dataset} & Musique & NQ & 2Wiki & HotpotQA & Bamboogle & StrategyQA & Avg. \\
    \midrule
    \textbf{Retrieval Counts} & & & & & & & \\
    \quad Exact Retrieval & 3.247 & 2.288 & 3.126 & 2.702  & 2.440 & 2.496 & 2.717 \\
    \quad \texttt{SearchAgent-X}       & 3.251 & 2.292 & 3.138 & 2.699 & 2.448 & 2.476 & 2.717 \\
    \midrule
    \textbf{Output Length} & & & & & & & \\
    \quad Exact Retrieval & 8125 & 5839 & 7575 & 6840 & 6152 & 6402 & 6822 \\
    \quad \texttt{SearchAgent-X}       & 8134 & 5847 & 7600 & 6839 & 6151 & 6382 & 6826 \\
    \bottomrule
  \end{tabular}
  \label{tab:SearchAgent-X-comparison-length}
\end{table}

\subsection{Detailed Overall Performance} \label{sec:appendix-overall-performance}
Table~\ref{tab:SearchAgent-X-all-metrics-2} presents more results of overall efficiency across different methods. We note that \texttt{SearchAgent-X} outperforms all baselines in different scenarios including model sizes, deployment methods (single GPU or distributed GPUs), and top-$k$ values. The advantage of \texttt{SearchAgent-X} is bigger in the 7B model and small top-$k$ values of the 14B model. This is due to two reasons. First, although the 14B model distributes model weights across two A100s, its KV-cache space is still smaller than the 7B model (because the model weights are larger, resulting in a larger KV-cache for a single token). The 7B model has a larger available KV-cache space, thus yielding greater benefits from managing the prefix cache. Second, the 14B model we test calls retrievals more cautiously, while the 7B model calls more retrievals (e.g., 4.9 of the 7B model vs. 3.3 of the 14B model when top-$k$ = 3 for Musique dataset), resulting in a greater distinction in request priority.

\subsection{Comparison of Different Priority Levels $G$} \label{sec:appendix-priority-level}
\begin{figure}
  \centering
  \includegraphics[width=\textwidth]{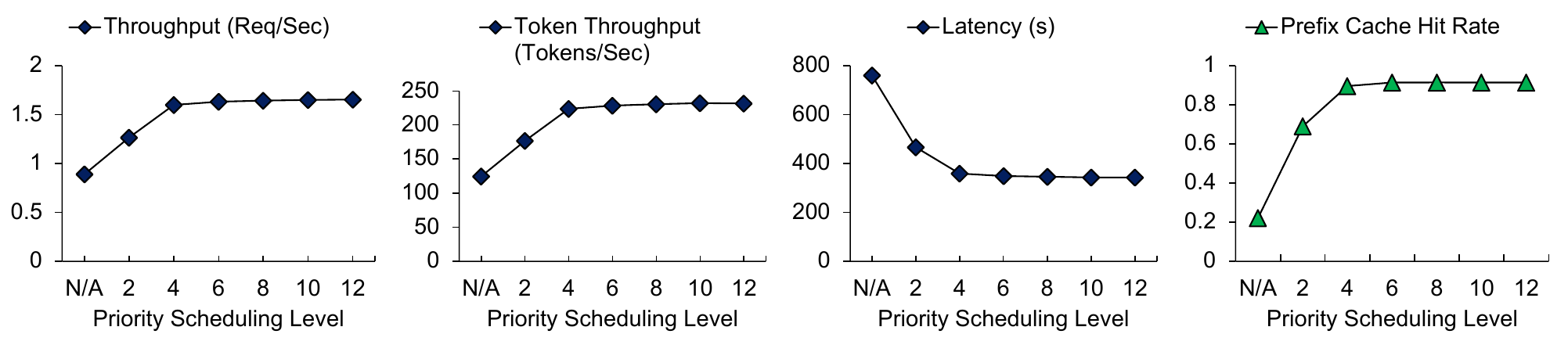}
  \caption{\textbf{Comparison of Different Priority Levels $G$.} {The numbers on the X-axis represent different priority scheduling levels; N/A indicates that priority scheduling is not used.}}
  \label{fig:SearchAgent-X_priority_level}
\end{figure}

\textbf{The performance of \texttt{SearchAgent-X} is insensitive to its priority level setting.}
The priority level $G$ mentioned in Section~\ref{sec:priority_scheduling} is an important hyperparameter of our method. In this section, we conduct an ablation study to evaluate the performance of \texttt{SearchAgent-X} with different priority levels $G$. We set $G$ = 2, 4, 6, 8, 10, and 12, and compare them with the baseline without priority scheduling (N/A). The results are shown in Figure~\ref{fig:SearchAgent-X_priority_level}.
We note that the performance of \texttt{SearchAgent-X} is not sensitive to the choice of $G$, and all efficiency metrics (including throughput, token throughput, latency, and prefix cache hit rate) first improve and then stabilize after $G$ = 6. This is expected because the average retrieval number of the 7B model is within 4 and 6, while the primary objective of priority scheduling is to distinguish requests with different retrieval numbers for effective management.

\subsection{Analysis of Maturity Exit Mechanism} \label{sec:appendix-analysis-maturity-exit}
\textbf{The maturity exit mechanism effectively halts unnecessary searches without compromising retrieval quality.}
To validate the effectiveness of non-stall retrieval, we analyze whether the maturity-based termination reliably halts unnecessary ANN iterations. We compare the retrieval traces of representative queries under two settings: maturity-based early stop and standard natural stop. As shown in Figure~\ref{fig:node-recall}, we make several observations. First, query difficulty varies significantly across the dataset, resulting in different numbers of candidate nodes explored before natural convergence. This highlights the need for an adaptive termination strategy rather than relying on a fixed exploration budget. Second, for queries of varying difficulty, the number of candidates explored by the maturity-based strategy closely matches the natural termination point of standard search, indicating our maturity exit accurately captures query difficulties. More importantly, the recall achieved by maturity-stopped queries remains consistently high (0.963 on average). These results confirm that our non-stall retrieval effectively terminates redundant search iterations while preserving retrieval quality.


\section{Limitations and Broader Impacts}\label{sec:appendix-limi}
While our method focuses on ANN-based dense retrieval—a widely adopted choice in modern LLM systems—it currently does not account for traditional keyword-based retrieval. In real-world scenarios, hybrid strategies that combine sparse and dense methods may offer complementary advantages. Extending our approach to support hybrid retrieval is a promising direction. Additionally, our current implementation uses a fixed dense encoder to transform text into vector representations. Although encoding latency contributes minimally to overall runtime, optimizing this stage may further improve retrieval quality.

On the positive side, the techniques developed in \texttt{SearchAgent-X} are generalizable and can be readily applied to other reasoning-centric models or iterative ANN-based pipelines. In the long term, this line of work may support practical search agents in real-world deployments, such as OpenAI’s DeepResearch~\cite{openai2024deepresearch} and xAI’s DeepSearch~\cite{xai2024grokagents}, thereby improving fair information access and generating economic value.

We are not aware of any negative societal impacts arising from this work.

\end{document}